\documentclass[10pt,twocolumn,letterpaper]{article}

\usepackage[pagenumbers]{cvpr} % To force page numbers, e.g. for an arXiv version

\usepackage{graphicx}
\usepackage{amsmath}
\usepackage{amssymb}
\usepackage{booktabs}
\usepackage{enumitem}

\usepackage[pagebackref,breaklinks,colorlinks]{hyperref}

\usepackage[capitalize]{cleveref}
\crefname{section}{Sec.}{Secs.}
\Crefname{section}{Section}{Sections}
\Crefname{table}{Table}{Tables}
\crefname{table}{Tab.}{Tabs.}

\def\cvprPaperID{5626} % *** Enter the CVPR Paper ID here
\def\confName{CVPR}
\def\confYear{2023}

\begin{document}

\title{Improving Interpretability via Regularization of Neural Activation Sensitivity}

\author{Ofir Moshe, Gil Fidel, Ron Bitton, Asaf Shabtai \\
Department of Software and Information Systems Engineering\\
Ben-Gurion University of the Negev
% \and
 }
\maketitle

%%%%%%%%% ABSTRACT
\begin{abstract}
State-of-the-art deep neural networks (DNNs) are highly effective at tackling many real-world tasks.
However, their wide adoption in mission-critical contexts is hampered by two major weaknesses - their susceptibility to adversarial attacks and their opaqueness. 
The former raises concerns about the security and generalization of DNNs in real-world conditions, whereas the latter impedes users' trust in their output. 
In this research, we (1) examine the effect of adversarial robustness on interpretability and (2) present a novel approach for improving the interpretability of DNNs that is based on regularization of neural activation sensitivity. 
We evaluate the interpretability of models trained using our method to that of standard models and models trained using state-of-the-art adversarial robustness techniques.
Our results show that adversarially robust models are superior to standard models and that models trained using our proposed method are even better than adversarially robust models in terms of interpretability. 
\end{abstract}

\begin{figure*}
    \centering\includegraphics[width=0.99\textwidth,trim={0 0 3cm 0},clip]{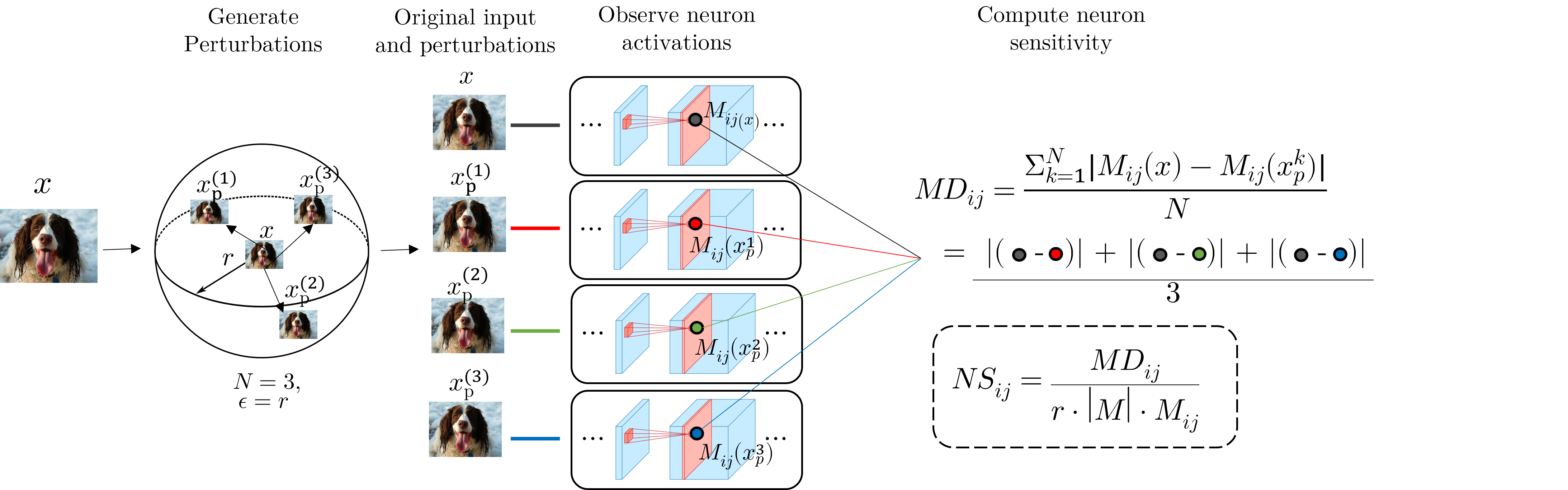}
    \caption{The process used to compute neuron activation differences (MD) and sensitivity (NS) for a single neuron $ij$ with a single example $x$. 
    A neuron's sensitivity is measured by its behavior, \ie, the variation in its activation values for a given input $x$ and the corresponding $N$ perturbations $x^k_p$ sampled from the surrounding $L_2$ sphere.
    $M_{ij}(\cdot)$ denotes the activation value of neuron $j$ in layer $i$ for a given input; each activation value is presented in a different color, \eg, the activation of the last perturbation, $x^3_p$, is presented in blue. 
    Then, $MD$ is used to calculate activation differences, and $NS$ is used to assess a neuron's sensitivity based on the computed $MD$ and normalization factors. 
    Note that this simplified example of NsLoss computation is presented to demonstrate the intuition and isn't a complete description of the computation process.}
    \label{fig:method}
\end{figure*}

%%%%%%%%% BODY TEXT
\section{\label{sec:intro}Introduction}

In recent years, deep neural networks (DNNs) have increasingly been used to tackle many complex tasks previously thought to be solvable only by humans, with accuracy often surpassing that of humans.
These tasks are part of technical domains such as computer vision, natural language processing, and anomaly detection, and use cases such as medical diagnosis, text translation, self-driving cars, fraud detection, and malware detection.

Two major obstacles to the wider adoption of DNNs in mission-critical tasks are (1) their vulnerability to adversarial attacks~\cite{goodfellow2014explaining,carlini2017adversarial,kurakin2016adversarial,chen2019hopskipjumpattack,brendel2017decision} and, more generally, concerns about their robustness when confronted with real-world data, and (2) their opaqueness, which makes it difficult to trust their output~\cite{the_need_for_xai_2021,ribeiro2016should}.

Extensive research has been performed to address the first obstacle, mainly consisting of approaches for the detection of adversarial examples~\cite{metzen2017detecting,feinman2017detecting,song2017pixeldefend,fidel2019explainability,katzir2018detecting} and methods for training robust models~\cite{goodfellow2014explaining,madry2017towards,salman2020adversarially,wong2020fast}. 
To address the second obstacle, research has focused on creating a priori interpretable models and developing methods for creating post-hoc explanations for existing models~\cite{smilkov2017smoothgrad,sundararajan2017axiomatic_integrated_gradients,shap2017,ribeiro2016should}.

The terms \textit{interpretability} and \textit{explainability} are often incorrectly used interchangeably. 
We adopt the definitions proposed by Gilpin \etal~\cite{gilpin2018explaining}:
\textit{Interpretability} is a measure of how well a human can understand the way a system (in our case - a DNN) functions. 
\textit{Explainability of DNNs} is a field of research with the goal of either answering the question of why a DNN produces a specific output for a given input (for example, by assigning contribution scores to different neurons of the network signifying their importance in steering the network towards that output) or describing what the network ``learns," \eg, producing visualizations of concepts learned by specific neurons~\cite{xai_feature_viz_olah2017feature}).

In this paper, we examine the relationship between the obstacles mentioned above by performing quantitative and qualitative analysis of the effect of a model's robustness (achieved via adversarial training and Jacobian regularization) on its interpretability, and by introducing a regularization-based approach which is conceptually similar to other approaches for improving adversarial robustness but has a substantial effect on the model's interpretability.

Recent research has shown that adversarial robustness positively affects interpretability~\cite{zhang2019interpreting,tsipras2018robustness,feature_purification_2022,noack2021empirical}. However, these studies mainly evaluated their methods using low resolution inputs and datasets such as the MNIST~\cite{lecun1998gradient} and CIFAR-10~\cite{krizhevsky2009learning}, and only provided anecdotal evidence of improved interpretability by presenting saliency maps of a few input images.
Moreover, these studies failed to pinpoint the specific trait of adversarially robust models that makes them more interpretable.

We hypothesize that the most important trait in this respect is the models' increased robustness to \textit{random} noise in a certain $\epsilon$-radius of the data manifold, as opposed to adversarial robustness which is essentially an approximation of a model's robustness to the \textit{worst possible} perturbations of a certain $\epsilon$-radius. 
Therefore, we construct a regularization term explicitly aimed at reducing the model's sensitivity to \textbf{random} perturbations.

Our main contributions in this paper are: (1) a quantitative comparison, based on well-accepted metrics, of the interpretability of adversarially trained models vs. standard ones on a high-resolution image dataset; (2) the discovery that Jacobian regularization is approximately as effective as adversarial training for improving model interpretability; (3) a novel, regularization based approach that outperforms both adversarial training and Jacobian regularization in the interpretability of the trained model, with a similarly low decline in accuracy.

\section{Background}
\label{sec:background}

\subsection{Post-hoc Explainability Techniques for DNNs}
There are many methods for explaining the predictions of machine learning models, and more specifically DNNs. 
In the computer vision domain, explanations usually consist of a saliency/attribution map which scores each input pixel based on its positive or negative effect on encouraging the model to classify the input as a specific class.
Some of the most prominent methods used in our evaluation include:
\begin{itemize}[noitemsep,topsep=0pt]
    \item \textbf{Integrated Gradients (IG)}~\cite{sundararajan2017axiomatic_integrated_gradients} calculates the importance score by summing the gradients on images interpolated between a baseline $x'$ and the input image $x$. 
    The baseline image $x'$ represents the absence of the input features. 
    Therefore, by computing the path integral between the baseline $x'$ and the real input $x$ of the partial derivative of the model over each feature, we obtain multiple estimates of the importance of every feature; this avoids the problem of saturated local gradients.
    Formally, the importance score for feature $i$ is computed as:
    \begin{equation}
        IG_i(f,x,x') = (x_i - x'_i)\times\int_{\alpha=0}^{1}\dfrac{\partial f(x'+\alpha(x-x'))}{\partial x_i}d\alpha
        \label{eq:ig}
    \end{equation}
    \noindent where $f$ represents the model, $x_i$ represents one feature of the input, and $\alpha$ is part of the integral and defines the distance on the path between $x$ and $x'$.
    \item \textbf{SHapley Additive exPlanations (SHAP)}~\cite{shap2017} (specifically \textit{Gradient SHAP}) is a local explainability method that approximates the Shapley values~\cite{shapley_values} of the input features by computing the expected values of the gradients when adding Gaussian noise to each input.
    Since it computes the expectations of gradients using different reference points, it can be viewed as an approximation of Integrated Gradients.
\end{itemize}

\subsection{Adversarial Evasion Attacks and Defenses}
Adversarial evasion attacks are methods for producing \textit{adversarial examples} -- model inputs that closely resemble valid inputs but result in drastically different model outputs. 
Formally, given a classifier $M(\cdot): R^d \rightarrow R^C$, an input sample $x \in R^d$, and a correct class label $c$, we call $\delta \in R^d$ an \textit{adversarial perturbation} and $x' = x + \delta$ an \textit{adversarial example} if:

\begin{equation}
    \begin{aligned}
        M(x') \neq c, \\
        s.t: ||\delta|| < \epsilon        
    \end{aligned}
\end{equation}
where $||\cdot||$ is a distance metric, and $\epsilon > 0$ is the maximum perturbation size allowed, which is set at a small positive value to constrain the perturbation so that the resulting adversarial example is indistinguishable from the original sample to the naked human eye, thus making it potentially useful in various adversarial scenarios.

Extensive research has been performed on countering adversarial attacks, mostly focused on methods for detecting adversarial examples and methods for training robust models. 
The latter is of greater relevance to the current research, and below we highlight two such methods:\\
\textbf{\textit{Adversarial training}}~\cite{goodfellow2014explaining,madry2017towards} is a method in which a model is trained to correctly classify adversarial examples by presenting them to the model during the training process. 
More precisely, this method solves the saddle point problem: $$\min_\theta \mathbb{E}_{x,y}\left[\max_{\delta: |\delta| < \epsilon}L(\theta, x + \delta, y)\right]$$ in which the aim is to find model parameters that minimize the expected value of the worst case increase in model loss due to input perturbations. 
Practically, the method consists of modifying the standard training loss so that it is applied to adversarial examples constructed from the training batch samples instead of the original training samples. 
Madry \etal~\cite{madry2017towards} used projected gradient descent (PGD) for generating adversarial examples during model training. \\
\textit{\textbf{Jacobian regularization}}~\cite{Jakubovitz_2018,hoffman2019robust} is a method in which the Frobenius norm of the Jacobian matrix containing the partial derivatives of the model's logits over the inputs is added as an extra loss term, resulting in the minimization of the Jacobian norm during model training. 
This was found to effectively push the model's decision boundaries away from the data manifold~\cite{hoffman2019robust} thus improving the model's robustness to adversarial attacks.

\section{\label{sec:related}Related Work}
The notion of adversarially robust models being more interpretable is not new. 
Zhang and Zhu~\cite{zhang2019interpreting} showed that adversarially trained convolutional neural networks (CNNs) produce explanations that rely on the global shape of the input images, in contrast to standard CNNs which focus more on textures that are inherently more sensitive to small perturbations.
Tispras~\etal~\cite{tsipras2018robustness} observed that adversarially trained models, by virtue of the constraints imposed by adversarial training that have the effect of reducing sensitivity to small perturbations, are more aligned with human vision, which is evident in explanations that emphasize features that are more human-perceivable.
In~\cite{feature_purification_2022}, the authors studied the effect of adversarial training on feature-level explanations of internal CNN layers and showed that adversarially trained models produce feature-level explanations that are ``purified" in that they are much less noisy and better represent high-level visual concepts. 
Noack~\etal~\cite{noack2021empirical} proposed a method for leveraging model explanations to improve robustness, by adding terms to the training loss that penalize the cosine of the angle between the explanation vector and the loss gradient, as well as the norm of the loss gradient vector. 
They showed that minimizing these two terms improves adversarial robustness.

\section{\label{sec:method}Proposed Method}

\subsection{Method Overview}
The proposed method aims to improve the interpretability of neural network classifiers. We introduce \textit{NsLoss}, a novel regularization term that penalizes the classifier for high sensitivity of the network's neurons to input perturbations.
Thus, we define a new training loss function as follows:
\begin{equation}
\label{eq:1}
\begin{aligned}
L = CELoss + \lambda \cdot NsLoss 
\end{aligned}
\end{equation} 
\noindent where $CELoss$ is the standard cross entropy loss, and $NsLoss$ is our new loss term.

We apply the proposed method on a pretrained model by continuing to train it with the custom loss function for a predefined number of epochs using a standard stochastic gradient descent-based optimizer and a relatively low learning rate to allow the model's interpretability to improve without ``unlearning'' the classification task.

\subsection{\label{subsec:nsloss_computation}NsLoss Regularization Term}

The inputs and parameters used to compute the NsLoss are as follows:
\begin{itemize}[leftmargin=*,noitemsep,topsep=0pt]
    \item $M$ - The model being trained.
    \item $X$ - Batch of samples for which the loss is computed.
    \item $\epsilon_{ns}$ - Hyperparameter specifying the radius of the $L_2$ ball in which random perturbations used to compute the loss are generated.
    \item $N$ - Number of perturbations to generate for each sample.
\end{itemize}
We begin by computing the normalized sensitivity of each neuron in the model to random perturbations of the input within an $L_2$ ball with radius $\epsilon_{ns}$. 
Given the $j$-th neuron of the $i$-th layer of the model:

\begin{enumerate}
    \item For every sample $x \in X$, generate $N$ random samples: $x_1, x_2, ... , x_N$ s.t. $||x-x_k|| = \epsilon_{ns}$ and store them in tensors $X_p^{(k)}$ for $k \in [1, N]$.
    \item Evaluate $M_{i,j}(X)$ and $M_{i,j}(X_p^{(k)})$, the activations of the $j$-th neuron in the $i$-th layer of the model on each original and each randomly perturbed sample, respectively.
    \item Compute the mean absolute activation of  neuron $i,j$ on the batch $X$: 
    \begin{equation}
        \label{eq:2}
        \begin{aligned}
            MA_{i,j} \gets \frac{\sum_{m=1}^{|X|}{|M_{i,j}(X[m])|}}{|X|}
        \end{aligned}
    \end{equation} 
    \noindent where $X[m]$ is the $m$-th sample in the batch $X$.
    \item Compute the mean absolute difference between the activations of the neuron on perturbed and original samples: 
    \begin{equation}
        \label{eq:3}
        \begin{aligned}
            MD_{i,j} \gets \frac{\sum_{k=1}^{N}\sum_{m=1}^{|X|}{|{M_{i,j}(X[m])-M_{i,j}(X_p^{(k)}[m])|}}}{N \cdot |X|}
        \end{aligned}
    \end{equation} 
    \item Compute the sensitivity of the neuron:
    \begin{equation}
        \label{eq:4}
        \begin{aligned}
            NS_{i,j} \gets \frac{MD_{i,j}}{\epsilon_{ns} \cdot |M_i| \cdot MA_{i,j}}
        \end{aligned}
    \end{equation} 
    \noindent where $|M_i|$ is the number of neurons in the $i$-th layer of $M$.
\end{enumerate}

Figure~\ref{fig:method} illustrates the process discussed so far, for a single neuron $ij$ and a single example $x$. 
 
\noindent
Finally, we compute the NsLoss as follows:
    \begin{equation}
        \label{eq:5}
        \begin{aligned}
            NsLoss(M, X, \epsilon_{ns}, N) \gets \sum_{i=1}^{|M|}\sum_{j=1}^{|M_i|}{NS_{i,j} \cdot MA_{i,j}}
        \end{aligned}
    \end{equation} 
    
The final loss is simply the mean neuron sensitivity weighted by the neuron's mean absolute activation, which accommodates for the neuron's contribution to the models' output. 
From the implementation perspective, it is important to note that although we described the algorithm for computing NsLoss so that we compute the neuron sensitivity values for each neuron separately (for simplicity's sake), in practice, it is straightforward to implement the computation of the aggregated loss using tensor operations on entire input batches and model layers, effectively making the time spent on loss computation negligible compared to model forward passes.

\subsection{The Effectiveness of NsLoss Regularization}
The NsLoss regularization term is constructed in a way that penalizes the model for small random input perturbations causing large differences in the activations of both the output neurons (logits) and internal neurons of the model.
This has the obvious effect of optimizing the model to minimize the activation differences and, as a result, to minimize the magnitude of the model's gradients with regard to inputs in the vicinity of the training set and, by generalization, the test set.
Once aggregated on the entire training set during the training process, this is expected to have the effect of minimizing the model's gradients in an $\epsilon_{ns}$ neighborhood of the entire data manifold, where $\epsilon_{ns}$ is the hyperparameter specifying the radius of the $L_2$ ball from which random input perturbations are sampled during the computation of the NsLoss regularization term.

\subsection{Hyperparameters}
A description of the hyperparameters used in our approach is provided below.
\begin{itemize}[noitemsep]
    \item \textbf{$N$} - The number of perturbed samples used to estimate the neuron sensitivity;  $N=5$ was used in all of our experiments.
    % This parameter has a substantial impact on the performance as additional N forward propagations of the model are performed for each input sample, increasing the training time in a linear fashion.
    \item \textbf{$\epsilon_{ns}$} - The radius of the $L_2$ ball from which perturbations for neuron sensitivity computation are generated.
    \item \textbf{$\lambda$} - The weight of the loss term. 
    We strive to choose the largest value of $\lambda$ that does not harm the model's cross-entropy loss and validation set accuracy.
    Specifically, we follow the protocol below to select the value of $\lambda$:
    \begin{enumerate}
        \label{lambda_protocol}
        \item For a standard model, compute the values of $NsLoss$ on 10 random batches from the training and validation sets and store the average value as $NsLoss_0$.
        \item Choose $\lambda_0 = \frac{\log_2{NumClasses}}{NsLoss_0}$.
        \item Perform a binary search by selecting values of $\lambda$ that are larger and smaller than $\lambda_{0}$ but in the same order of magnitude. 
        For each such $\lambda$, start training the model for an epoch. 
        If the training cross-entropy loss reaches the cross-entropy of a random guess ($\log_2{NumClasses}$), then $\lambda$ is too high; otherwise it can be increased further. 
    \end{enumerate}
\end{itemize}
\begin{figure*}
    \includegraphics[width=\textwidth,height=13cm]{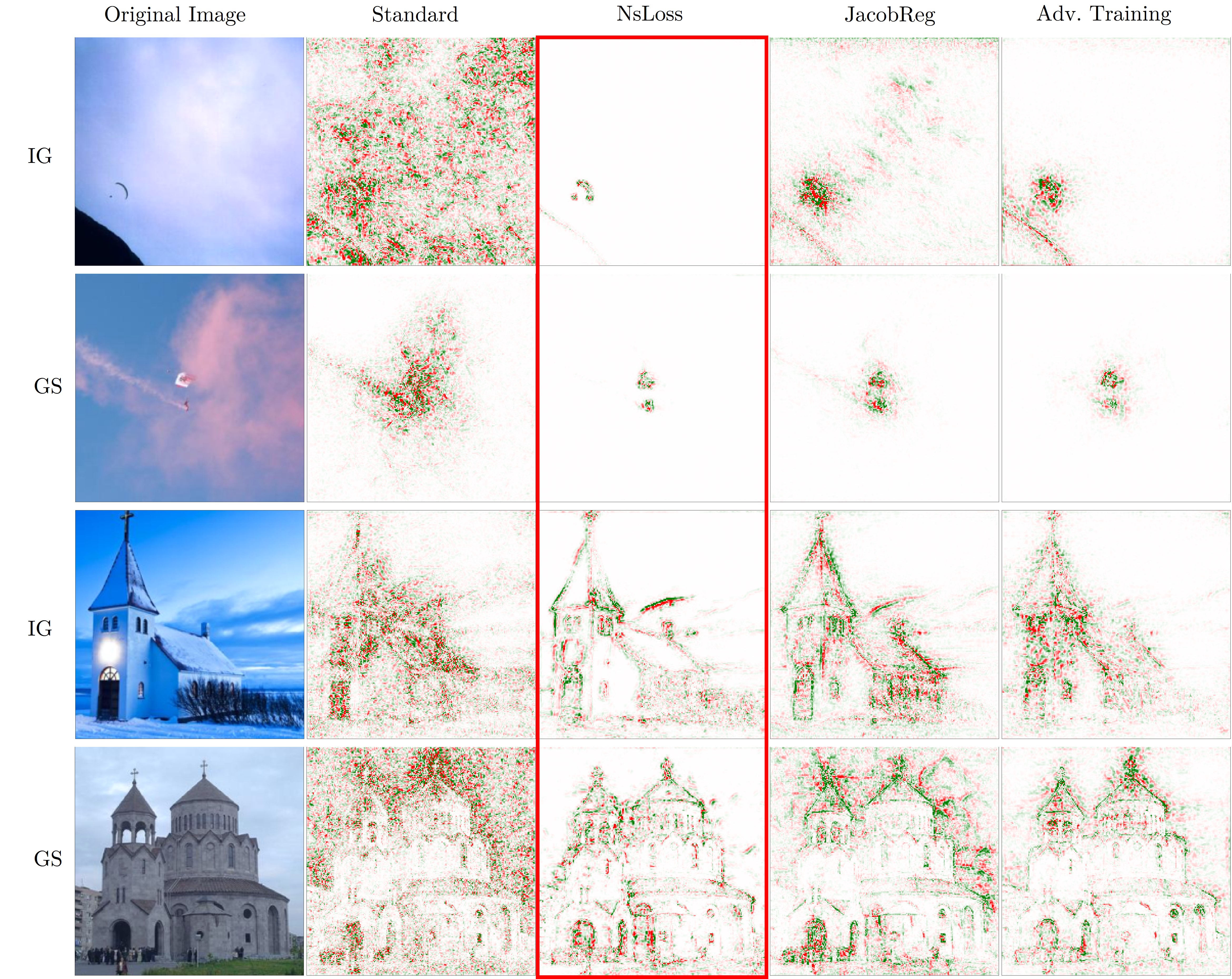}
      \caption{Two images of a parachute (top rows) and two images of a church (bottom rows), as well as a comparison of the attribution maps obtained for these images by the VGG19 model~\cite{simonyan2014very} trained using the standard, NsLoss, JacobReg and adversarial training methods. 
      The labels on the left indicates the attribution method used (IG and GS).
      The attribution maps generated by our method are framed in red.}
  \label{fig:vgg19compare_vis}
\end{figure*}
\begin{figure*}
    \includegraphics[width=\textwidth,height=7cm]{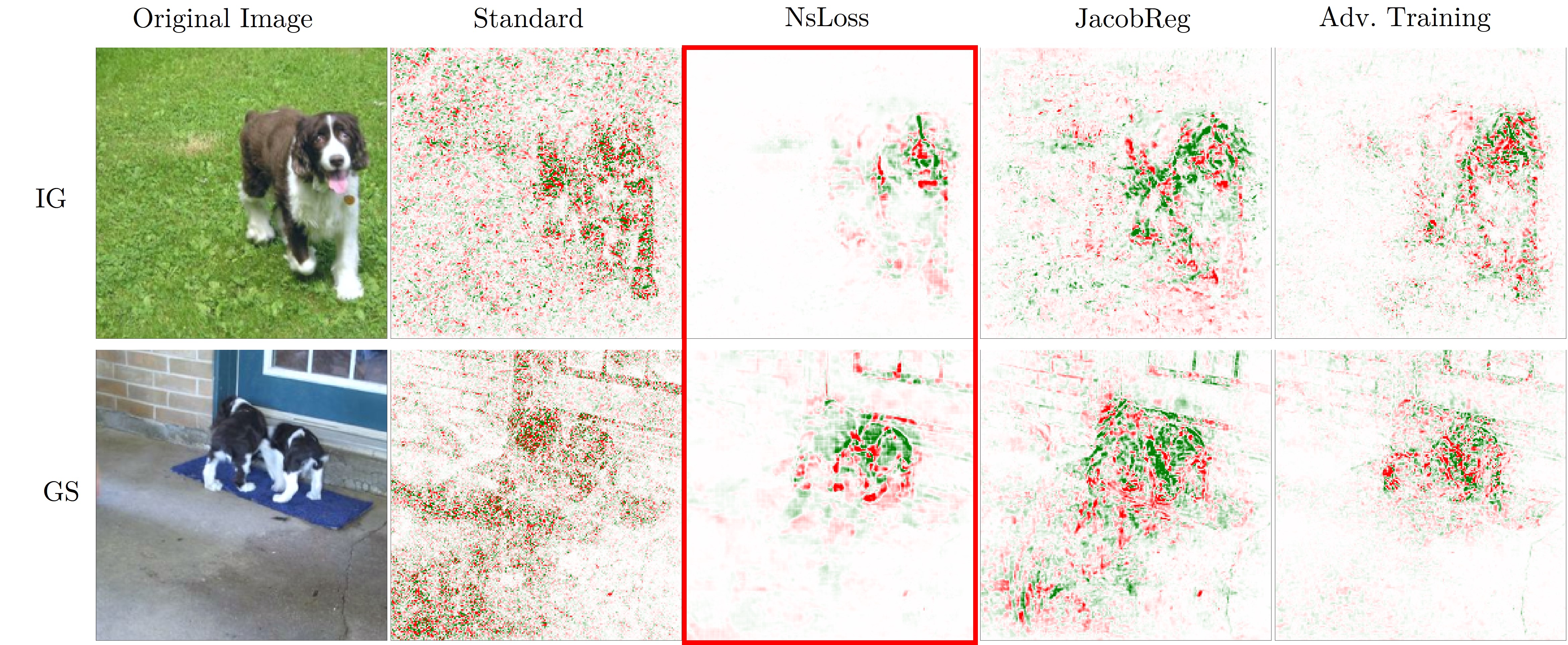}
      \caption{Comparison of the feature maps obtained by the PreResNet10~\cite{he2016preresnet} model which was trained using the standard, NsLoss, JacobReg, and adversarial training methods on images of dogs from the ImageNette~\cite{imagenette} test set.
    The labels on the left indicates the attribution method used (IG and GS).
    The attribution maps generated by our method are framed in red. }
  \label{fig:presnet_compare_vis}
\end{figure*}
\begin{table*}
\centering
\resizebox{450pt}{!}{
\begin{tabular}{@{}cccc|c|cc|ccc|cc@{}}
\toprule
              & \multicolumn{3}{c|}{General Training} &           & \multicolumn{2}{c|}{NsLoss params} & \multicolumn{3}{c|}{PGD params} & \multicolumn{2}{c}{Additional Info} \\
              & method        & epochs     & lr       & $\lambda$ & $N$        & $\epsilon$        & $D$        & $\epsilon$ & steps & Clean Accuracy    & Robust Accuracy   \\ \midrule
Standard      & fine-tune     & 5          & 1e-2     & -         & -          & -                 & -          & -          & -     & 0.968         & 0.00          \\
NsLoss        & retrain      & 80         & 1e-4     & 100       & 5          & 20                & -          & -          & -     & 0.900         & 0.49          \\
JacobReg      & retrain      & 80         & 1e-4     & 0.8       & -          & -                 & -          & -          & -     & 0.902         & 0.47          \\
Adv. Training & retrain      & 40         & 5e-4     & -         & -          & -                 & $L_\infty$ & $8/255$    & 7     & 0.792         & 0.58          \\ \bottomrule
\end{tabular}
}
\caption{Configurations of the PreResNet10 models.}
\label{tab:preresnet_conf}
\end{table*}
\begin{table*}
\centering
\resizebox{450pt}{!}{
\begin{tabular}{@{}cccc|c|cc|ccc|cc@{}}
\toprule
              & \multicolumn{3}{c|}{General Training} &           & \multicolumn{2}{c|}{NsLoss params} & \multicolumn{3}{c|}{PGD params} & \multicolumn{2}{c}{Additional Info} \\
              & method        & epochs     & lr       & $\lambda$ & $N$        & $\epsilon$        & $D$        & $\epsilon$ & steps & Clean Accuracy    & Robust Accuracy   \\ \midrule
Standard      & fine-tune     & 5          & 1e-2     & -         & -          & -                 & -          & -          & -     & 0.962         & 0.02          \\
NsLoss        & retrain      & 120        & 1e-4     & 9.5       & 5          & 20                & -          & -          & -     & 0.931         & 0.13          \\
JacobReg      & retrain      & 70         & 1e-4     & 0.05      & -          & -                 & -          & -          & -     & 0.929         & 0.15          \\
Adv. Training & r-train      & 50         & 1e-4     & -         & -          & -                 & $L_\infty$ & $4/255$    & 7     & 0.941         & 0.42          \\ \bottomrule
\end{tabular}
}
\caption{Configurations of the VGG19 models.}
\label{tab:vgg19_conf}
\end{table*}
\section{Evaluation}
\label{sec:eval}
The objective of our experiments is to examine the effect of adversarial training, Jacobian regularization, and our proposed NsLoss regularization term on the quality of the explanations both qualitatively (\ie, visual improvement) and quantitatively (using objective metrics). 
We also compare the results to those obtained by a baseline model.

\subsection{Datasets, Models, Robust Training Methods}
In all of our experiments we use the ImageNette~\cite{imagenette} dataset, which contains high resolution images from a 10 class subset of the popular ImageNet~\cite{deng2009imagenet} dataset.
We use VGG19~\cite{simonyan2014very} and PreResNet10~\cite{he2016preresnet} models pretrained on ImageNet and fine-tuned on ImageNette for a clean test accuracy of 96.2\%, and 96.8\% respectively. 
We use Madry's adversarial training method~\cite{madry2017towards}, as implemented by the ``robustness" library~\cite{robustness_library} to train robust models for our evaluation.
We retrain the model for 50 epochs against a PGD adversary in the $L_\infty$ norm, using seven PGD steps, $\epsilon=4/255$ and with random initialization.
We use the implementation of Jacobian regularization presented by Hoffman~\etal~\cite{hoffman2019robust}.
Tables~\ref{tab:preresnet_conf} and~\ref{tab:vgg19_conf} summarize the configurations and hyperparameters used to train all of the models evaluated.

\begin{table*}[]
\centering
\resizebox{450pt}{!}{%
\begin{tabular}{cccccccc|c|c}
&
  \multicolumn{3}{c|}{\textbf{Sensitivity}} &
  \multicolumn{2}{c|}{\textbf{Faithfulness}} &
  \multicolumn{2}{c|}{\textbf{Complexity}} &
  \multicolumn{2}{c}{\textbf{Accuracy}} \\
& Max ($\downarrow$) & Avg ($\downarrow$) & \multicolumn{1}{c|}{LL ($\downarrow$)} & Corr ($\uparrow$) & \multicolumn{1}{c|}{Est ($\uparrow$)} & Com ($\downarrow$)   & Spr ($\uparrow$)  & \multicolumn{1}{l|}{Clean Accuracy} & Robust Accuracy         \\ \hline
\textbf{Standard}             & 0.0233             & 0.0226             & 52.95                                & -0.02           & -0.22                               & 10.19              & 0.59            & 0.97                           & 0.00                 \\
\textbf{NsLoss}               & \underline{0.0002}             & \underline{0.0002}             & \textbf{10.32}                         & \textbf{0.22}     & \textbf{0.46}                         & \textbf{9.89}        & \textbf{0.68}     & 0.90                            & 0.49                 \\
\textbf{JacobReg}             & \underline{0.0002}             & \underline{0.0002}             & 20.21                                  & 0.06              & \underline{0.15}                                  & 10.02                & 0.65              & 0.90                            & 0.47                 \\
\textbf{Adv. Training}        & \textbf{0.0001}    & \textbf{0.0001}    & \underline{19.89}                                  & 0.15              & \underline{0.42}                                  & \underline{9.99}                 & \underline{0.64}              & 0.79                            & 0.58                 \\ \hline
\end{tabular}%
}

\caption{Comparison of attribution quality on the PreResNet10 model trained using the four different methods presented in~Table~\ref{tab:preresnet_conf}; Integrated Gradients \cite{integratedgradients} is used as the base attribution method. 
($\uparrow$) indicates that a higher value is better, and ($\downarrow$) indicates that a lower is  better. A value in \textbf{bold} indicates the best score in the column, whereas an \underline{underlined} value indicates that this value is second best.
The scores were computed and averaged on the entire ImageNette test set.}
\label{tab:preresnet}
\end{table*}
\begin{table*}[h]
\centering
\resizebox{450pt}{!}{%
\begin{tabular}{@{}cccccccc|cc@{}}
 &
  \multicolumn{3}{c|}{\textbf{Sensitivity}} &
  \multicolumn{2}{c|}{\textbf{Faithfulness}} &
  \multicolumn{2}{c|}{\textbf{Complexity}} &
  \multicolumn{2}{c}{\textbf{Accuracy}} \\
 &
  Max ($\downarrow$) &
  Avg ($\downarrow$) &
  \multicolumn{1}{c|}{LL ($\downarrow$)} &
  Corr ($\uparrow$) &
  \multicolumn{1}{c|}{Est ($\uparrow$)} &
  Com ($\downarrow$) &
  Spr ($\uparrow$) &
  Clean Accuracy. &
  Robust Accuracy \\ \midrule
\textbf{Standard} &
  0.635 &
  0.556 &
  43.521 &
  -0.021 &
  -0.176 &
  10.043 &
  0.638 &
  0.962 &
  0.02 \\
\textbf{NsLoss} &
  {\underline{0.004}} &
  {\underline{0.004}} &
  \textbf{14.955} &
  {\underline{-0.007}} &
  \textbf{-0.076} &
  \textbf{9.634} &
  \textbf{0.744} &
  0.931 &
  0.13 \\
\textbf{JacobReg} &
  \textbf{0.002} &
  \textbf{0.002} &
  21.891 &
  \textbf{0.000} &
  {\underline{-0.133}} &
  9.944 &
  0.665 &
  0.929 &
  0.15 \\
\textbf{Adv. Training} &
  0.060 &
  0.058 &
  {\underline{16.654}} &
  -0.031 &
  -0.222 &
  {\underline{9.824}} &
  {\underline{0.694}} &
  0.941 &
  0.42 \\ \bottomrule
\end{tabular}%
}
\caption{Comparison of attribution quality on the VGG19 model trained using the four different methods presented in Table~\ref{tab:vgg19_conf}; IntegratedGradients~\cite{integratedgradients} is used as the base attribution method.
($\uparrow$) indicates that a higher value is better, and ($\downarrow$) indicates that a lower value is better. 
The scores were computed and averaged on the entire ImageNette test set.}
\label{tab:my-table}
\end{table*}

\subsection{Evaluation Metrics} 
Various recent studies~\cite{carvalho2019machine,montavon2018methods,alvarez2018robustness} attempted to determine what properties an attribution-based explanation should have. 
They showed that one metric alone is insufficient to provide explanations that are meaningful to humans.
As suggested by Bhatt \etal~\cite{bhatt2020evaluating}, three desirable criteria for feature-based explanation functions are: low \emph{sensitivity}, high \emph{faithfulness}, and low \emph{complexity}.
Therefore, we evaluate the different techniques based on these three well-studied properties:

\begin{enumerate}[noitemsep,topsep=0pt]
\item {\em \textbf{Sensitivity}} - measures how strongly the explanations vary within a small local neighborhood of the input when the model prediction remains approximately the same~\cite{alvarez2018robustness,yeh2019fidelity}. 
In our evaluation, we use max-sensitivity, avg-sensitivity~\cite{yeh2019fidelity}, and the local Lipschitz estimate~\cite{alvarez2018robustness}.

\item {\em \textbf{Faithfulness}} - estimates how the presence (or absence) of features influences the prediction score; \ie, whether removing highly important features results in model accuracy degradation~\cite{bhatt2020evaluating,bach2015pixel,alvarez2018towards}. 
In our evaluation, we use the faithfulness correlation~\cite{bhatt2020evaluating} and faithfulness estimate~\cite{alvarez2018towards}.

\item {\em \textbf{Complexity}} - captures the complexity of explanations \ie, how many features are used to explain a model's prediction~\cite{chalasani2020concise,bhatt2020evaluating}. 
In our evaluation, we use complexity~\cite{bhatt2020evaluating} and sparseness~\cite{chalasani2020concise}.
\end{enumerate}

\subsection{Explainability Methods}
We use the Integrated Gradients~\cite{sundararajan2017axiomatic_integrated_gradients} and Gradient SHAP~\cite{shap2017} explanation methods in our evaluation.

We argue that since the majority of local explanation methods use model gradients, any improvement on these explanation methods using the proposed method is likely to be successfully transferred to other gradient-based methods.

\subsection{Quantitative Evaluation Results}
We start by examining the performance of the compared methods, considering the three aforementioned explanation-quality criteria (\ie, sensitivity, faithfulness, and complexity), applied to the values of the methods' respective explanations.

The results are summarized in Table~\ref{tab:preresnet} for the PreResNet10 model and Table~\ref{tab:my-table} for the VGG19 model. 
In both Tables, the methods (standard, NsLoss, JacobReg, and adversarial training) are listed in the first column, and the respective values for the metrics: max-sensitivity (Max), avg-sensitivity (Avg)~\cite{yeh2019fidelity}, local Lipschitz estimate (LL)~\cite{alvarez2018robustness}, faithfulness correlation (Corr)~\cite{bhatt2020evaluating}, faithfulness estimate (Est)~\cite{alvarez2018towards}, complexity (Comp)~\cite{bhatt2020evaluating} and sparseness (Spr)~\cite{chalasani2020concise}, are presented in columns 2-8. 

For the sensitivity criteria, lower values are better; for the faithfulness criteria, higher values are better, and for the complexity criteria, lower complexity values and higher sparseness values are better. 
The scores were computed and averaged over the entire test set from the ImageNette dataset. 
The Quantus library~\cite{quantus} was employed for XAI evaluation and Integrated Gradients was used as the base attribution method. 
To facilitate meaningful comparisons, all models compared were retrained to have roughly the same test accuracy on natural images (except for the standard model), which is presented in column 9. 
To support the hypothesis that robust models tend to be more interpretable, we also provide the robust accuracy for each method, obtained using the AutoAttack~\cite{autoattack} library, in column 10.  

\begin{figure}
\includegraphics[width=\columnwidth,height=7cm]{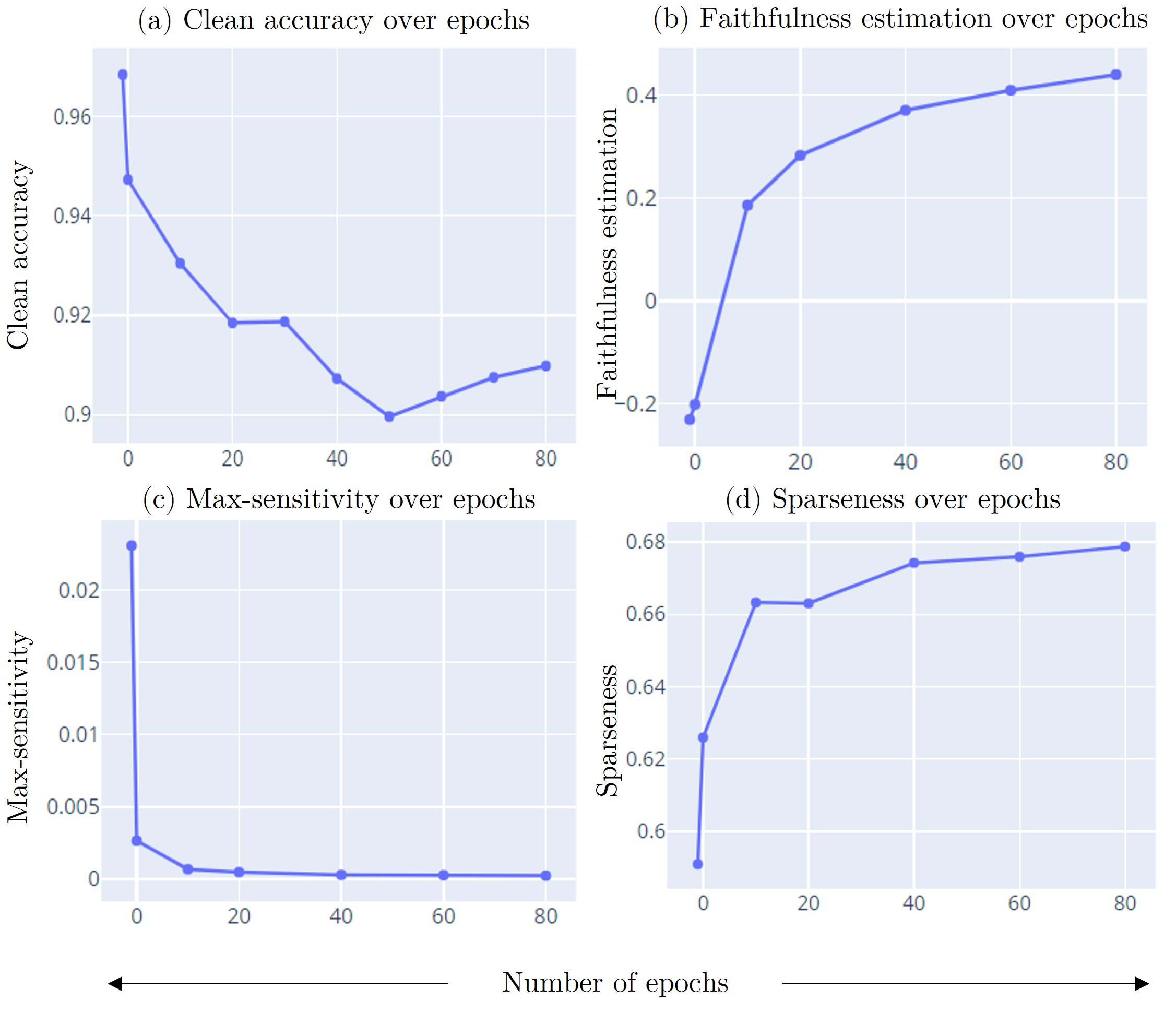}
  \caption{The effect of the number of training epochs on test accuracy and a subset of explanation-quality metrics. 
  The scores were obtained by training PreResNet10 for 80 epochs using the NsLoss method with hyperparameter $\lambda=60$. 
  In (a) clean accuracy over epochs, (b) faithfulness estimation~\cite{alvarez2018towards} over epochs, (c) max-sensitivity~\cite{yeh2019fidelity} over epochs, (d) sparseness~\cite{chalasani2020concise} over epochs.}
  \label{fig:over_epochs}
\end{figure}

\begin{table}[]
\centering
\resizebox{\columnwidth}{!}{%
\begin{tabular}{@{}ccccc@{}}
\toprule
\textbf{$\lambda$} & \textbf{Max-Sensitivity}~\cite{yeh2019fidelity} & \textbf{Faithfulness Est.}~\cite{alvarez2018towards} & \textbf{Sparseness}~\cite{chalasani2020concise} & \textbf{Accuracy} \\ \midrule
\textbf{60}        & 0.0016                   & 0.15                      & 0.66                & 0.93              \\
\textbf{100}       & 0.0012                   & 0.26                      & 0.66                & 0.92              \\
\textbf{200}       & 0.0008                   & 0.38                      & 0.69                & 0.90              \\
\textbf{500}       & 0.0005                   & 0.43                      & 0.70                & 0.86              \\
\textbf{1000}      & 0.0003                   & 0.45                      & 0.71                & 0.78              \\ \bottomrule
\end{tabular}%
}
\caption{The effect of the NsLoss $\lambda$ hyperparameter (rows) on the accuracy and the explanation metric (columns). 
The scores were obtained when PreResNet10 was trained for 10 epochs with the corresponding $\lambda$ value.}
\label{tab:lambda_effect}
\end{table}

In Table~\ref{tab:preresnet}, we examine the PreResNet10 model trained using the four different methods presented in Table~\ref{tab:preresnet_conf}.
Compared to the other methods, our proposed NsLoss method results in a significant improvement in attribution quality, and it achieves the best performance on nearly all of the metrics. 

With regard to the sensitivity criterion, NsLoss and JacobReg obtain a comparable score in the max-sensitivity and avg-sensitivity metrics, which represents a reduction of over 116\% in sensitivity compared to the standard method, while adversarial training leads, obtaining a score that is twice as good as theirs but this comes at a cost of a 10\% decrease in the clean test accuracy.

On the  local Lipschitz
estimate metric, NsLoss is superior with an improvement of over 500\% vs. the standard method; this is followed by adversarial training with an improvement of 266\%.
With regard to the faithfulness criteria, NsLoss outperforms the other methods by a significant margin on both metrics. 
Furthermore, for the complexity criterion, the NsLoss method achieves the best scores on the two metrics examined.

Table~\ref{tab:my-table} presents a similar performance comparison, this time using variants of the VGG19 model presented in Table~\ref{tab:vgg19_conf}.
Regarding the sensitivity criterion, the NsLoss and JacobReg methods significantly improve the sensitivity, with a decrease of 317\% and 156\% respectively for both the max and avg sensitivity metrics, whereas for the local Lipschitz
estimate metric NsLoss outperforms both alternatives.
For the faithfulness criterion, on the faithfulness correlation metric, the JacobReg method is clearly superior, whereas on the faithfulness estimation metric, NsLoss performed favorably.
For the complexity criterion, NsLoss performed the best in all metrics tested, albeit by a moderate margin.

Based on the quantitative evaluation results, we can conclude with sufficient certainty that (1) robust models are superior to standard models in terms of interpretability, since their explanations are less sensitive, more faithful, and less complex; and (2) the proposed NsLoss method produces models that are more interpretable than those produced by both the adversarial training and Jacobian regularization methods.

\subsection{Qualitative Evaluation Results}
Figures~\ref{fig:vgg19compare_vis} and~\ref{fig:presnet_compare_vis} 
present the attribution maps of images from ImageNette's test set for VGG19 and PreResNet10 respectively, 

The NsLoss attribution maps for the parachute images (top two rows in Figure~\ref{fig:vgg19compare_vis}) demonstrate the  ability of the trained NsLoss model to capture the key region of interest in the image (the parachute itself) and produce an attribution map that focuses precisely on that region, while ignoring the background.
The other methods compared (except for the standard method) were also able to capture the region of interest in the image but these regions were rather noisy much less sharp.

Following Smilkov \etal~\cite{smilkov2017smoothgrad}, we use \emph{visual coherence} to indicate that the salient areas highlight mainly the object of interest, rather than the background.
As can be seen in the attribution maps of the church images (bottom two rows in Figure~\ref{fig:vgg19compare_vis}) and the dogs (in Figure~\ref{fig:presnet_compare_vis}), the standard saliency maps demonstrate quite poor visual coherence, as they focus mainly on the background, rather than on the object itself. 
In contrast to this, the NsLoss, JacobReg, and adversarial training methods provide more visually coherent maps; however, based on the figures provided, as well as the comprehensive analysis we performed, NsLoss is found to consistently provide the most visually coherent and least noisy maps compared to the other methods, regardless of the explainability method used.

\subsection{Effect of Regularization Term  ($\lambda$) and Training Length}
NsLoss makes use of several hyperparameters. 
We present the effect of (1) $\lambda$, the weight of the regularization term, and (2) the number of training epochs on select interpretability metrics.
\paragraph{Regularization term weight ($\lambda$)}
In Table~\ref{tab:lambda_effect}, it can be observed that as $\lambda$ increases, there is a gradual improvement in the results for all explanation-quality metrics: The max-sensitivity scores decrease and the faithfulness estimation and sparseness increase, suggesting an improvement of the models' interpretability. 
However, this improvement is accompanied by a certain drop in the model's accuracy. 
Therefore, the $\lambda$ value must be carefully chosen; we suggest that readers follow the protocol described in Section~\ref{lambda_protocol}.

\paragraph{Training epochs}
Figure~\ref{fig:over_epochs} presents plots for three chosen attribution quality metrics and clean test accuracy, over 80 training epochs. 
It can be seen that the max-sensitivity values decrease relatively quickly, right from the first epoch. Both the faithfulness estimation and sparseness continue to improve moderately as epochs progress. 
Moreover, the results show that there is an interpretability-accuracy trade-off, and a gradual drop in accuracy can be seen. Therefore, the training process should be monitored, choosing the `sweet spot` where there is a balance between the desired explanation quality and the required accuracy of the model.

\section{\label{sec:conclusion}Conclusions and Future Work}
Our experimental results validate the effectiveness of both adversarial training, Jacobian regularization, and our novel regularization-based approach (NsLoss) in improving the models' interpretability by changing the model's behavior such that state-of-the-art explainability methods produce explanations that are more focused and better aligned with human perception.
This supports previous research results and hypotheses about the positive effect of adversarial training on model interpretability and sets the stage for further research in the field. 
Moreover, we quantitatively demonstrated the superiority of our proposed method using well-accepted metrics for measuring the quality of explanations, as well as representative qualitative evidence based on saliency map visualizations.

Future work may include: (1) testing our method on other computer vision tasks. 
We expect to see very similar results on other datasets, with medical imaging being a natural choice, since it is a field where interpretability is crucial in order to establish trust in the system's output; (2) applying NsLoss to other domains (beyond computer vision), which should be fairly straightforward as the method makes no assumptions about the nature of the input or the model's architecture; and (3) explore the effect of NsLoss regularization on the nature of features learnt by the model, similar to related work conducted for adversarially trained models~\cite{feature_purification_2022,zhang2019interpreting,tsipras2018robustness}. 
Extrapolating the results presented in this paper leads us to believe that NsLoss trained models learn features that are even more aligned with human perception than adversarially trained models.

%%%%%%%%% REFERENCES
\newpage
{\small
\bibliographystyle{ieee_fullname}
\bibliography{egbib}
}

\end{document}